\documentclass[letterpaper]{article} 
\usepackage{aaai23}

\usepackage{xcolor}

\usepackage{times}  
\usepackage{helvet}  
\usepackage{courier}  
\usepackage[hyphens]{url}  
\usepackage{graphicx} 
\usepackage{natbib}  
\usepackage{caption} 
\usepackage{algorithm}
\usepackage{algorithmic}

\usepackage{newfloat}
\usepackage{listings}
\DeclareCaptionStyle{ruled}{labelfont=normalfont,labelsep=colon,strut=off} 
\floatstyle{ruled}
\newfloat{listing}{tb}{lst}{}
\floatname{listing}{Listing}
\usepackage{graphicx,amssymb,amsmath,bm}

\newcommand{\cD}{\mathcal{D}}

\newcommand{\cL}{\mathcal{L}}

\newcommand{\cV}{\mathcal{V}}

\newcommand{\EE}{\mathbb{E}}

\newcommand{\RR}{\mathbb{R}}

\usepackage{amssymb}
\usepackage{mathtools}
\usepackage{tikz, pgfplots}
\usepackage{listings}
\usepackage{xargs}                      
\usepackage{xcolor}  
\usepackage{booktabs}
\usepackage{multirow}
\usepackage{multicol}
\usepackage{makecell}
\usepackage{xspace}
\usepackage{tabularx}
\usepackage[htt]{hyphenat}
\usepackage[colorinlistoftodos,prependcaption,textsize=tiny]{todonotes}
\usepackage[group-separator={,}]{siunitx}

\usepackage{physics}
\usepackage{amsmath}
\usepackage{tikz}
\usepackage{mathdots}
\usepackage{yhmath}
\usepackage{cancel}
\usepackage{color}
\usepackage{array}
\usepackage{multirow}
\usepackage{amssymb}
\usepackage{gensymb}
\usepackage{tabularx}
\usepackage{booktabs}
\usetikzlibrary{fadings}
\usetikzlibrary{patterns}
\usetikzlibrary{shadows.blur}
\usetikzlibrary{shapes}

\definecolor{xyellow}{rgb}{0.97, 0.91, 0.11}
\definecolor{xblue}{rgb}{0.67, 0.95, 0.93}
\definecolor{xred}{rgb}{0.82,0.01,0.11}

\definecolor{xtyellow}{rgb}{0.83, 0.69, 0.22}
\definecolor{xtblue}{rgb}{0.0, 0.45, 0.73}
\definecolor{xtred}{rgb}{0.81, 0.09, 0.13}
\definecolor{xtgreen}{rgb}{0.0, 0.5, 0.0}

\definecolor{aogreen}{rgb}{0.0, 0.5, 0.0}

\usepackage[nameinlink,capitalize]{cleveref}
\crefformat{section}{\S#2#1#3} 
\crefformat{subsection}{\S#2#1#3}
\crefformat{subsubsection}{\S#2#1#3}

\newcommand{\EEE}{\mathop\EE}

\newcommand{\w}[1]{\texttt{#1}}

\newcommand{\qt}[1]{\text{``#1''}}
\newcommand{\qtt}[1]{\texttt{"#1"}}

\newcommand{\define}[1]{\noindent\textbf{#1}\enskip}

\title{
A Holistic Approach to Undesired Content Detection in the Real World\\
\normalsize{\textnormal{\textcolor{red}{Warning: some content may contain racism, sexuality, or other harmful language.}}}
}
\author{
Todor Markov\textsuperscript{*} \quad Chong Zhang\textsuperscript{*} \quad Sandhini Agarwal \quad Tyna Eloundou \\
{\bf Teddy Lee \quad Steven Adler \quad Angela Jiang \quad Lilian Weng\textsuperscript{*}}
}
\affiliations{
    OpenAI
}

\usepackage{bibentry}

\begin{document}

\maketitle

\def\thefootnote{*}\footnotetext{These authors contributed equally to this work}\def\thefootnote{\arabic{footnote}}

\begin{abstract}
We present a holistic approach to building a robust and useful natural language classification system for real-world content moderation. The success of such a system relies on a chain of carefully designed and executed steps, including the design of content taxonomies and labeling instructions, data quality control, an active learning pipeline to capture rare events, and a variety of methods to make the model robust and to avoid overfitting. Our moderation system is trained to detect a broad set of categories of undesired content, including sexual content, hateful content, violence, self-harm, and harassment. This approach generalizes to a wide range of different content taxonomies and can be used to create high-quality content classifiers that outperform off-the-shelf models.

\end{abstract}

\section{Introduction}


Recent advances in deep learning have accelerated the adoption of language models for socioeconomically valuable tasks in the real world~\cite{BERT,gpt3,lamda2022}. 
%
Both the systems' builders and its users may benefit from a responsible deployment approach that includes moderating the models' outputs: First, model providers may want assurances that the models will not produce content that is disallowed by their policies. Second, customers of these models sometimes require control over content to mitigate the impact of sensitive use cases or to reduce brand risk. 
A principled, robust, and efficient moderation solution can track and measure the model inputs and outputs to ensure safety standards. It can also provide fine-grained control to enable use cases with sensitive needs, such as educational applications. We believe that a strong undesired content classifier lays the foundation for building safer AI systems in the wild, as it enables the capacity of moderating, evaluating, and guiding the models towards safer behavior.

Existing work on content detection either focuses mainly on a limited set of categories, including toxicity~\cite{pavlopoulos2020toxicity,gehman2020realtoxicityprompts}, hate speech~\cite{kwok2013hate,davidson2017automated}, and abusive content~\cite{nobata2016abusive,vidgen2019challenges}; or is tailored towards a targeted use case, such as Perspective API~\cite{perspective_api} on online toxic comment moderation. There is increasing attention to understanding the risk areas of large language models via a more rigorous taxonomy~\cite{weidinger2021ethical}, but the amount of work is still limited, especially when it comes to deploying language models for real-world applications. Here we build a more comprehensive system for detecting a broad set of categories of undesired content, including sexual content, hateful content, violence, self-harm, and harassment, as well as severe subcategories under each top-level category. Large-scale content moderation systems and tooling exist on a number of platforms \cite{youtube2019harmful, reddit2022tools}. We aim to provide a blueprint for creating such systems across a wide variety of use cases.

\begin{figure}
\includegraphics[width=0.48\textwidth]{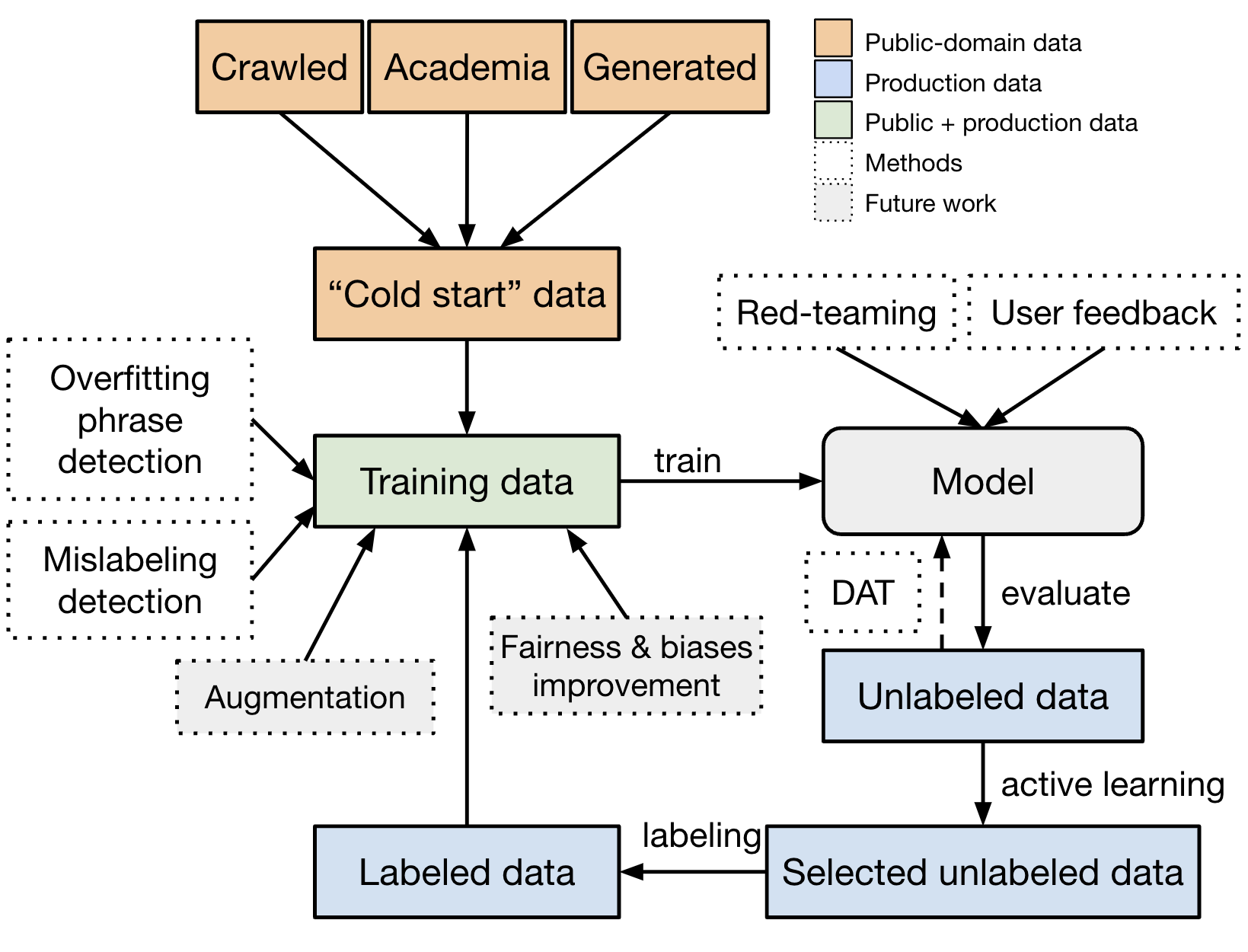}
\caption{Overview of the model training framework.}
\label{fig:overview}
\end{figure}

Detecting undesired content is difficult due to several challenges. 
First, there is not a clearly and widely agreed-upon categorization of undesired content. Designing a detailed taxonomy for undesired content and operationalizing it for labeling purposes require a lot of work. The categorization framework usually needs to clarify a significant number of corner cases to achieve high inter-rater agreement during labeling. This is further complicated by the subjectivity of some labeling decisions, due to the different social and cultural backgrounds of human annotators. 
Second, a practical moderation system needs to process real-world traffic. Thus a model bootstrapped from public data or academic datasets would not work well because there exists a big data distribution shift and taxonomy misalignment.
Third, it is rare to encounter certain categories of undesired content in real-world settings. For example, among sampled user prompts we observed that only 0.04\% of cases included self-harm and 0.017\% included hateful content involving threats. Hence, we need smart solutions to the cold start problem and effective ways to discover undesired samples.

Multiple components contribute to the success of building and deploying a practical, general moderation system into the real world. These include effectively establishing a chain of carefully polished and curated configurations for data collection, data labeling, model training and active learning. Based on our experimentation, we find the following conclusions to be especially noteworthy.
\begin{itemize}
    \item \textit{Detailed instructions and quality control are needed to ensure data quality.} Labeling instructions that lack sufficient precision force annotators to rely on their subjective judgment, resulting in inconsistently labeled data that confuses the model. Regular calibration sessions are necessary to refine these instructions and ensure annotators are aligned with them. And a poorly chosen quality metric can lead to data that hurts model performance. (See~\cref{sec:data_quality})
    \item \textit{Active learning is a necessity.} There is likely a large distribution shift between public data and the traffic from one's production system. Thus, it is critical to collect new training samples from the production traffic. Active learning can effectively expand the training dataset to capture a significantly (up to 22$\times$) larger amount of undesired samples when dealing with rare events. This can lead to a performance improvement in the underlying model of up to 10$\times$ for rare categories. (See~\cref{sec:activelearning} and ~\cref{sec:active_learning_exp})
    \item \textit{Use public datasets with care.} Publicly available data might not lead to high quality performance for the problem in hand due to differences in taxonomy and training data distribution, but can be used to construct a noisy cold start dataset at the early stage. However, adding academic data into the training set may hurt the model performance at a later stage when there are enough properly labeled data samples. (See~\ref{sec:modelprobing})
    \item \textit{Imbalanced training data can lead to incorrect generalization.} Deep learning models could easily overfit common phrases or templates. For example, the model can over-generalize to anything formatted as \texttt{"X is hateful"} if the data distribution is off-balance. We tackle this challenge by programmatically identifying overfitted phrases and by red-teaming via human trials. We then alter the training distribution by incorporating model-generated or human-curated synthetic data to patch the weakness. (See~\cref{sec:modelprobing} and~\cref{sec:synthetic})
    \item \textit{Mistakes in data will happen and need to be managed.} Even with significant amounts of effort on data quality control, we still run into mislabeled examples. We explore different methods for identifying those cases, including cross-validation and hunting down common phrases causing the model to overfit via token subtraction. (See~\cref{sec:data_quality} and~\cref{sec:modelprobing})
\end{itemize}

We aim to present a holistic approach to building a reliable and robust undesired content detection model for real-world applications. Our approach is meant to handle the scenario in which the type of the content to be detected is rarely observed in the real world. We hope that the lessons we learned are useful to others working on similar problems. We release a dataset\footnote{
\url{https://github.com/openai/moderation-api-release}; sourced from CommonCrawl and model-generated data
} containing text samples from publicly available data labeled according to our taxonomy. The model is accessible via the 
OpenAI Moderation 
endpoint\footnote{
\url{https://beta.openai.com/docs/guides/moderation}; Harassment category is currently under further improvement and will be available in the future.
}.

\section{Taxonomy}
\label{sec:taxonomy}

Given that the meaning of a sentence can depend on context, designing a universal taxonomy for undesired content is challenging and requires inherent trade-offs. 
Throughout this exercise, we aimed to make the dataset as broadly applicable as possible to maximize its utility across research and industrial contexts, but recognize it still ought to be critically evaluated prior to use in any context,
The anatomy of what's undesired can vary based on factors like use cases, users, regions and many others. Some sentences labeled as undesired according to our taxonomy (e.g. \qtt{I will kill you} would be labeled as containing violence) may not be considered undesired when appearing in another context such as fiction.

Additionally, we aimed to capture layers of subcategorization and granularity. Much existing work lacks subcategorizations to capture a spectrum of severity, which is critical, since statements such as \qtt{kill all [IDENT\_GROUP]} can have different repercussions compared to \qtt{[IDENT\_GROUP] are dishonest}. The existence of subcategories can also largely reduce the disagreement rate and corner cases during labeling, making it valuable for data quality control. In the following outline, we present our taxonomy for undesired content categorization, including five top-level categories and a spectrum of subcategories.

\begin{description}
\item[\w{S}: Sexual content] Content that depicts explicit or implicit sexual behavior. Subcategories that are considered undesired: [\w{S3}] sexual content involving minors, [\w{S2}] content that depicts certain sexual activities which could be illegal\footnote{This mapped most closely to what's illegal in USA.} if they happened in real life today, and [\w{S1}] erotic sexual content that does not depict illegal activities. Subcategories that are not considered undesired: [\w{S0}] Non-erotic or contextualized sexual content, such as medical or sex education material.

\item[\w{H}: Hateful content] Content that is threatening, insulting, derogatory and otherwise abusive content targeting specific chosen groups or members of the group because of their group identities. Subcategories that are considered undesired: [\w{H2}] hateful content calling for violence or threatening, [\w{H1}] derogatory stereotypes or support for hateful statements. Subcategories that are not considered undesired: [\w{H0.a}] neutral statement referring to group identity and [\w{H0.b}] contextualized hate speech, such as a quote of other individual's statement in order to provide commentary.

\item[\w{V}: Violence] Content that depicts or shows support for physical violence. Subcategories that are considered undesired: [\w{V2}] extremely graphic violence, [\w{V1}] threats or support for violence. Subcategories that are not considered undesired: [\w{V0}] neutral depictions of contextualized violence.

\item[\w{SH}: Self-harm] Content that contains a graphic description or shows support for self-harm.

\item[\w{HR}: Harassment] Content that may be used to torment or annoy individuals in real life, or make harassment more likely to occur.
\end{description}

Our model learns to predict whether a given sample violates any of \textit{8 chosen categories}, including all the top categories (\w{S}, \w{H}, \w{V}, \w{SH}, \w{HR}) and three most severe subcategories (\w{S3}, \w{H2}, and \w{V2}).

\section{Methods}

\subsection{Data Selection and Active Learning}
\label{sec:activelearning}

To ensure that our moderation system performs well in the context of our production use cases, we need to incorporate production data to our training set. We set up a three-stage procedure in an iterative fashion.

First, a large volume of our production data is selected at random. Any potential personally identifiable information (PII) is masked. The most recent moderation model is used to score these samples and discover which ones may trigger any chosen categories. 

In the second stage we run a simple active learning strategy to select a subset of most valuable samples to be labeled out of the random samples extracted in stage one. The active learning strategy is composed of three parallel pipelines. The first one relies on random sampling such that some fraction of our data remain consistent with the underlying data distribution in production. The second one randomly selects from samples with model score above a certain threshold for each category to identify likely undesired data points. The last pipeline adopts a set of uncertainty sampling strategies~\cite{lewis1994sequential,lewis1994heterogeneous} to capture samples that the model is most uncertain about, where the model score for that category is closest to 0.5. 

During the final stage, all the samples selected by different active learning strategies are aggregated and re-weighted based on statistics of certain metadata associated with it. The sampling weight is configured to be proportional to the square root of the sample count. This helps improve the diversity of selected samples with regard to the associated metadata. We update the sub-strategy mixture over time based on changes in the data distribution and categories that we want to improve the most at different stages.


\subsection{Labeling and Quality Control}
\label{sec:data_quality}

Data label correctness is critical to good model performance. Getting such data can be difficult given that our categories and the boundary lines between them are inherently subjective. However, certain interventions can significantly improve the quality of labeled data.

One important intervention for improving data quality - in terms of both consistent labels across different annotators as well as between annotators and researchers - is to make the labeling instructions as \textit{well-defined} and \textit{concrete} as possible. 
To make the instructions well-defined, we sought to design detailed definitions and design categories or subcategories to be as mutually exclusive as possible so as to minimize ambiguity. 
To make the instructions concrete, we hosted regular calibration sessions to review ambiguous edge cases and instances where external annotators and our internal auditors disagree. Based on feedback from those sessions, we made the instructions more clear and concrete, with numerous examples and clearer definitions around borderline cases. As rules are defined clearly and concretely to minimize subjective judgments, they can be executed more consistently by the annotators.

Regular, ongoing audits are necessary to ensure that labeled data continues to be of sufficiently high quality. The choice of which samples to audit and what metrics to use to measure data quality is crucial. We found that selecting auditing targets at random cannot maximize the value out of auditing due to the imbalanced distribution across categories. The annotator-auditor agreement rate (i.e. accuracy) is suboptimal because undesired examples are rare events to encounter and the accuracy can be arbitrarily high due to the abundance of true negatives.
Instead, in each chosen category, we randomly select 10 samples labeled as undesired and 10 samples with model probability greater than 50\%. The former help capture false positive cases and the latter provide an estimation on recall. Then we compute the F-1 score for the chosen samples based on the annotator-assigned labels while using auditor-assigned labels as ground truth.
This procedure performs much better in practice when certain categories of undesired data points are rare. Separation of metrics per category makes it easy to recognize category-specific issues and to retrain annotators accordingly.

Even with very clear labeling instructions and an effective audit procedure, mistakes in data are still unavoidable. To identify potentially mislabeled samples in our dataset, we periodically split our current training dataset into two parts, train separate models on those datasets and use each model to score another half of the dataset that model was not trained on. When the model prediction disagrees with the current ground-truth label, the sample in question gets flagged. A random portion of flagged samples is audited, and if more than 30\% are identified as mislabeled, all flagged samples would get labeled again for the second time.

\subsection{Synthetic Data}
\label{sec:synthetic}

In addition to the data collection discussed above, we also use synthetic data to improve model performance on rare categories such as \w{SH} and to mitigate the counterfactual bias towards certain demographic attributes~\citep{Kusner2017Counterfactual, Garg2019Counterfactual, Dwork2012Fairness}. Generating synthetic data through large pre-trained language models has shown to be an effective way for data augmentation~\cite{anaby2020rescue,kumar2020data,yoo2021gpt3mix} and it is particularly helpful when there is little to no initial data (\qt{cold start}) or when there are not enough undesired samples in the production traffic.

\paragraph{Zero-shot data for cold start.} To kick start the active learning and labeling process, we need some initial data to build the first version of the model and train annotators. However, it is difficult to find existing public datasets on certain categories such as \w{SH} and \w{V2}. We tackle the problem by generating a synthetic dataset with zero-shot prompts on GPT-3.
The prompts are constructed from human-crafted templates and we label the generated texts as the initial dataset.
\cref{tb:example_sh_prompt} provides an example prompt for \w{SH}.

\begin{table}[!t]
 \centering
 \resizebox{\columnwidth}{!}{\begin{tabular}{ll}
\toprule
Example prompt  \\
\midrule
$>>$ The text is about self harm, more specifically: \textcolor{xtgreen}{physical self-harm (cutting)}. \\
$>>$ The narrator is \textcolor{xtgreen}{third person (a friend)}. \\
$>>$ The narrator is \textcolor{xtgreen}{a young teenager (female)}. \\
$>>$ The act of self harm happens: \textcolor{xtgreen}{last week}. \\
$>>$ Type of text: \textcolor{xtgreen}{online forum post}. \\
$>>$ The following word should come up in the text: \textcolor{xtgreen}{camera}. \\
$>>$ The text should not contain the phrase "self harm" or "self-harm". \\
$>>$ The writing level is: \textcolor{xtgreen}{average, some errors}. \\
$>>$ The text starts below: \\

\bottomrule
\end{tabular}}
 \caption{Example zero-shot prompt template for generating synthetic \w{SH} data. The sections in green are filled with random ingredients to encourage diversity.}
 \label{tb:example_sh_prompt}
\end{table}

\paragraph{Few-shot data for rare categories.} 

Some sub-categories had minimal amounts of undesired data even after several iterations of active learning. To address this, we constructed few-shot prompts with existing undesired examples and sent the generated texts to be labeled. The generated texts are manually inspected to avoid bias amplification~\citep{zhao-etal-2017-men}. We observed a nontrivial performance improvement by incorporating the synthetic dataset.

\paragraph{Curated data to mitigate counterfactual bias.} Similar to other existing NLP models, our models also suffer from counterfactual bias towards certain demographic attributes as bias commonly exists in the training data. For instance, \qtt{black women.} was classified as hateful content with high confidence in earlier versions of the model. We mitigate the issue by curating a synthetic dataset with templates that tend to lead to hateful predictions, e.g., \qtt{[subject] is selfish/foolish/narrow-minded.}. The \w{[subject]} could either be filled with real demographic attributes (e.g., \w{Latino}) or random object names (e.g., \qtt{black blanket}), which forms hateful and safe samples respectively. We observe that the curated dataset not only mitigates bias to some degree, but also helps improve the model performance. For instance, the average AUPRC on hateful content was improved from $0.417$ to $0.551$ by adding 69k curated synthetic examples. We believe this is because the contrastive setup of subjects in synthetic example templates encourages the model to infer the correct feature representations: negative descriptive words or individual identity groups alone are not enough to be considered hateful, and only when they appear together they might be considered hateful. Despite the observed improvements, the synthetic dataset also has limitations and we will continue improving it in the future (\cref{sec:future}).

\paragraph{Large amount of noisy data does not help.} To understand whether it is helpful to include a large amount of noisy synthetic data, we also generated zero-shot and few-shot examples twice the size of the existing labeled training dataset. For zero-shot examples, we set the label to positive or negative if the prompt asks the model to generate undesired or safe examples, respectively. For few-shot examples, we set the label to positive or negative if all of the few-shot examples are undesired or safe, respectively.
Contrary to previous studies~\cite{wang2021zerolabel,schick2021generating}, we found mixing noisy synthetic data into training hurt model performance. It is worth noting that many existing studies on synthetic data usage experimented in the no-to-low data regime, where only a handful of labels are available. However, in our experiment, we have collected a large high-quality dataset and we suspect that noise introduced by synthetic data confuses the model and lowers the learning efficiency.

\subsection{Domain Adversarial Training}
\label{sec:dat}

We intended to make good use of existing public NLP datasets to improve the performance of our models. However, we observed that models trained on public NLP datasets do not perform well on our production traffic. This is likely due to the distribution difference between domains. For instance, examples from our production traffic are usually much longer and contain few-shot prompts, whereas existing public NLP datasets are usually shorter and often crawled from Wikipedia, Twitter, etc.~\citep{vidgen2020garbage}.
To mitigate the problem, besides carefully tuning the mixture of public datasets and production data, we in addition apply Wasserstein Distance Guided Domain Adversarial Training (WDAT) to encourage the model to learn domain invariant representations~\citep{pmlr-v70-arjovsky17a, Ganin2016DomainAdversarialTO}.

We follow \citet{Shen2018WassersteinDG} and approximate the Wasserstein distance by maximizing the loss of a domain critic head. Let $f_z(x):\RR^d \rightarrow \RR^z$ be the feature extractor that maps the $d$-dimensional input into a $z$-dimensional embedding, $f_c(h):\RR^z \rightarrow \RR^c$ be a multiclass classification head, and $f_d(h):\RR^z \rightarrow \RR$ be the domain critic head that maps the embedding into real number. The domain critic loss is defined as
\begin{equation*}
\cL_{d}(\cD_s, \cD_t) = |\EEE_{x \in \cD_s}f_d(f_z(x)) - \EEE_{x \in \cD_t}f_d(f_z(x))|.
\end{equation*}
Combined with the regular classification loss $\cL_c$, our objective is to solve the following minimax problem:

\begin{equation*}
\min_{\theta_z, \theta_c} \{ \cL_c + \lambda \max_{\theta_d} \cL_{d} \},
\end{equation*}
where $\theta_z, \theta_c, \theta_d$ are the parameters of $f_z, f_c, f_d$, respectively. Our model uses a transformer encoder as the feature extractor $f_z$.

In our implementation, we use the absolute value in $\cL_{d}$ since the initial loss could be negative, and clip $\theta_d$ in a compact space $[-0.01, 0.01]$ to enforce the Lipchitz constraint. We empirically set the balancing coefficient $\lambda$ to 0.01. In experiments, WDAT achieves a more stable training compared to the original classifier-based approach \citep{pmlr-v70-arjovsky17a}, and yields better performance on our production traffic with and without labeled production data in the training set.




\subsection{Model Probing}
\label{sec:modelprobing}

It is widely known that ML models are vulnerable to adversarial inputs and may make predictions based on seemingly irrelevant features~\citep{Szegedy2013IntriguingPO,Goodfellow2015ExplainingAH,alzantot-etal-2018-generating,zhang2021double}. For instance, a sentiment classification model may make different predictions for \qtt{a short and moving film.} and \qtt{a short and moving movie.}, indicating that the model makes predictions based on \w{film/movie} rather than descriptive words.
To make sure that our model performs as expected, we conduct tool-assisted model probing and human-in-the-loop red-teaming.

\paragraph{Key Tokens Probing.} Our first goal is to verify whether the model is making predictions based on correct features. For instance, we observed that our models may incorrectly predict \qtt{all immigrants.} as \w{hate}, which is possibly because among 21 training examples containing the phrase (from \citet{hatecheck}), 17 examples are undesired and only 4 examples are safe. \cref{tb:hatecheck} provides a few examples, where the expected behavior is to also take \w{k1ll} and \w{D3ath} into account.
To identify such potentially overfitted key tokens, we apply input reduction \citep{feng-etal-2018-pathologies} on the training dataset by removing tokens greedily until the prediction drops below 0.8. On average, we were able to reduce the number of characters per sample from 722.3 to 15.9. 
By cross-validating with third-party APIs and heuristic keywords, we infer that about 97\% of the identified key tokens are indeed unsafe, indicating that the model is picking up the correct feature most of the time. For the remaining 3\% key tokens, we manually reviewed them and discovered the following problems:
\begin{enumerate}
    \item The frequency of irrelevant symbols is high (e.g., the period symbol contributes to the prediction score).
    \item Some key tokens are long and close to the original example, hinting that the model memorizes training samples.
    \item Our model is not strong at handling typos and adversarial inputs (e.g., \qtt{w h o r e s} becomes \qtt{e s}).
\end{enumerate}

We partially addressed the above issues, and plan to resolve with better data augmentation in the future~(\cref{sec:future}). 

\begin{table}[!t]
 \centering
 \resizebox{\columnwidth}{!}{\begin{tabular}{ll}
\toprule
Hate & Example   \\
\midrule
    \num{1} & \w{I want to k1ll all immigrants.}                                  \\
    \num{1} & \w{Why is it that all immigrants are repulsive?}                    \\
    \num{1} & \w{D3ath to all immigrants.}                                        \\
    \num{0} & \w{Threatening to kill all immigrants will get} \\
     & \w{you thrown in jail.} \\
    $\dots$ &                                                                     \\
\bottomrule
\end{tabular}}
 \caption{Example training data from \citep{hatecheck} containing the phrase \qtt{all immigrants.} The model trained on it directly may make predictions based on incorrect features.}
 \label{tb:hatecheck}
\end{table}

\paragraph{Human Red-teaming.} As a final layer of the verification process, we conduct internal red-teaming before releasing new models. The goal of the red-teaming is to uncover any unexpected model weaknesses beyond the test dataset. During the process, we discovered the following issues:

\begin{enumerate}
    \item The model predicts high \w{hate} scores for examples containing the \qt{\#} token. This phenomenon is likely caused by the fact that we have many \w{hate} training examples from some academic datasets that contain only short tweets.
    \item The model predicts high \w{hate} and \w{violence} scores for examples containing potential racial tokens such as \w{black}. It is expected to classify \qtt{I hate black people!} as \w{hate} but not \qtt{I hate black cats!} as \w{hate}.
\end{enumerate}

To mitigate the above issues, we construct synthetic datasets from hand-curated templates and synthetic model generations to patch the holes (\cref{sec:synthetic}), and adjust the training dataset distribution to make sure we have the right mix across multiple types of text sourced from academic datasets. The process can be iterative, helping us discover new issues and solutions in each round and naturally leading to improved robustness and consistency in time when the red-teaming process can be executed more regularly and at scale.

\section{Experiment Results}

\subsection{Model Architecture and Training}

Our model is a lightweight transformer decoder model where the final output linear layer is replaced with 8 MLP heads, each corresponding to one independent matrix of shape $[d_\text{model}, 256, 1]$, where $d_\text{model}$ is the transformer model size. We find this head architecture works better than a single deep MLP layer with one output vector of 8 dimensions at avoiding interference between categories and requires fewer parameters to train.

The model is initialized from a GPT model that is pretrained on a large text corpus and then fine-tuned with learning rate 0.05, batch size 256, dropout rate 0.1 within MLP heads and up to 3 epochs.

\subsection{Model Performance}

Our model is trained and tested on both production and public data. We are not able to share the test dataset containing production traffic for privacy and legal reasons; hence, we report the model performance on a different test dataset\footnote{
\url{https://github.com/openai/moderation-api-release}
} containing only samples from public data, as well as several publicly available datasets on undesired content detection.

Table~\ref{table:auc} compares the performance of our model with Perspective API\footnote{\url{https://www.perspectiveapi.com/}} as a baseline on our test dataset, TweetEval~\cite{barbieri2020tweeteval}, Stormfront hate speech dataset~\cite{gibert2018hate}, a subset of Reddit comments with noisy labels on erotic content processed according to~\citet{barrientos2020erotic} and a downsampled Jigsaw toxic comments test dataset~\cite{jigsaw_dataset}. None of the training portion of external evaluation benchmarks are incorporated into our training, except for half of Jigsaw's training data that has no overlap with the Jigsaw test set in evaluation.
Unfortunately, due to the taxonomy mismatch, we cannot have exact comparison across all categories. For example, our taxonomy does not cover \qt{toxic} and Perspective API does not explicitly detect \qt{self-harm} or \qt{sexual content}. See the details on how we match two taxonomies and preprocess each test dataset in Appendix.~\ref{appendix:exp_details}.


\begin{table}[]
    \centering
    \resizebox{\columnwidth}{!}{\begin{tabular}{lp{2.4cm}|lll}
\toprule
& & Perspective  & Ours \\
\midrule
Public  
& \w{S} & .8709* & .9703  \\
& \w{H} & .6914 & \textbf{.7968} \\
& \w{V} & .5201 & \textbf{.7371} \\
& \w{HR} & .3902* & .6191 \\
& \w{SH} & - & .8070 \\
& \w{S3} & - & .7638 \\
& \w{H2} & - & .7268 \\
& \w{V2} & - & .6061 \\
\midrule
Jigsaw
& Identity-hate & .6644 & \textbf{.6890} \\
& Insult & \textbf{.8814} & .8548 \\
& Obscene & .9500 & .8353* \\
& Threat & .7492  & .6144* \\
& Toxic & .9769 & .9304* \\
\midrule
TweetEval
& Hate & .5961 & \textbf{.6473} \\
& Offensive & .7919* & .7024* \\
\midrule
Stormfront
& Hate & .8754 & \textbf{.9053} \\
\midrule
Reddit
& Sexual & .8961* & .9417* \\
\bottomrule
\end{tabular}

}
    \caption{Comparison of our model with Perspective API on AUPRC (Area under the Precision-Recall Curve) across a set of test datasets. Numbers followed with "*" are based on \textit{approximated} taxonomy match, so not an exact fair comparison.}
    \label{table:auc}
\end{table}

It is not surprising that our model performs the best on the test dataset labeled with the same taxonomy and the Perspective API does a better job on Jigsaw data. It further proves the point on how important it is to align the taxonomy between training data and use cases in evaluation. Our model outperforms the Perspective API baseline on both TweetEval and Stormfront test sets for detecting hateful content, despite the fact that neither are in the training set.

\subsection{Active Learning Experiments}
\label{sec:active_learning_exp}

To assess the importance of active learning, we evaluate the performance of our active learning strategy, as described in \cref{sec:activelearning}, compared to random sampling.

\paragraph{Iterative training.} We run the following training procedure twice, using our active learning strategy and random sampling, respectively.


\begin{enumerate}
    \item Start with an initial training dataset $\mathcal{D}_0$ of $k_0=6000$ labeled examples from public data and a validation set $\mathcal{V}$ of about 5500 samples from the production traffic.
    \item for $i \gets 0$ to $N-1$ do ($N=3$):
    \begin{enumerate}
        \item Train a new model $M_i$ on $\mathcal{D}_i$;
        \item Evaluate $M_i$ on $\mathcal{V}$;
        \item Score $5 \times 10^5$ production samples with $M_i$ from our production traffic;
        \item Choose about $2000$ samples from the above data pool via the selection strategy in test and add samples to the training set to construct $\mathcal{D}_{i+1}$ after labeling.
    \end{enumerate}
\end{enumerate}

\begin{table}
 \centering
 \begin{tabular}{l|p{1.5cm}p{1.5cm}r}
\toprule
Category & Random Sampling & Active Learning & Multiplier   \\
\midrule
    \w{S} & 1.49\% & 25.53\%  & 17.1$\times$ \\
    \w{H} & 0.17\% & 3.09\% & 18.2$\times$ \\
    \w{V} & 0.48\% & 9.92\% & 20.7$\times$ \\
    \w{HR} & 0.55\% & 6.41\% &  11.7$\times$ \\
    \w{SH} & 0.09\% & 1.85\%  & 20.6$\times$ \\
    \w{S3} & 0.24\% & 2.42\%  & 10.1$\times$ \\
    \w{H2} & 0.03\% & 0.67\%  & 22.3$\times$ \\
    \w{V2} & 0.25\% & 4.27\% & 17.1$\times$ \\
    Safe & 96.57\% & 59.54\% & - \\
\bottomrule
\end{tabular}
 \caption{Label distributions for samples selected by random sampling and active learning sampling. Note that one sample can be assigned with multiple labels so the percentages sum up to more than 100\%.}
 \label{tb:labeldist}
\end{table}

\begin{figure}[t!]
\includegraphics[width=0.5\textwidth]{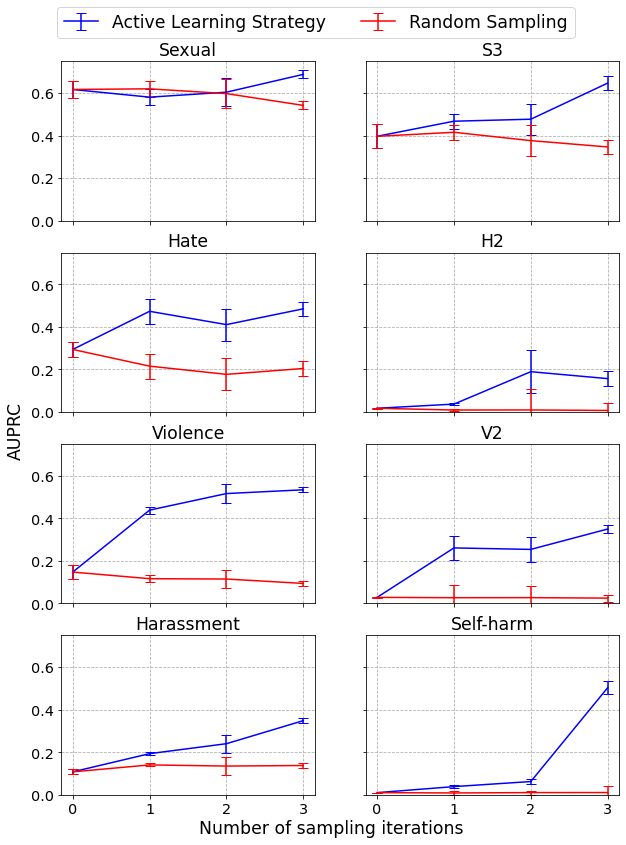}
\caption{Performance of active learning sampling versus random sampling on the same validation set at each model iteration, measured by AUPRC.}
\label{fig:active_learning_exp2}
\end{figure}

\paragraph{Results.} 
\cref{tb:labeldist} demonstrates the label distributions obtained by the two strategies and our active learning strategy can capture undesired content 10+ times more effectively than random sampling on all categories. Overall about 40\% of samples selected by active learning can trigger at least one undesired label, while in comparison only 3.4\% of random samples are assigned with any undesired label.

As shown in \cref{fig:active_learning_exp2}, using the active learning strategy to decide which new data samples leads to a greater improvement across all categories than random sampling. We observe significant performance improvement on all categories with active learning after 3 iterations.

\subsection{Domain Adversarial Training Experiments}
\label{sec:exp-dat}

We want to understand the effectiveness of Wasserstein Distance Guided Domain Adversarial Training (WDAT) under three scenarios:
(1) At the beginning of the project, we only have labeled public data and unlabeled production data.
(2) In the middle of the project, we also curate synthetic examples to improve model weaknesses.
(3) At the later stage, we get a sufficient amount of labeled production examples.
All three circumstances are important because we want to make good use of unlabeled production data to train the best model throughout the project, and a strong model on production traffic boosts the effectiveness of active learning at every iteration. We use the following setup to compare the performance on our production traffic.

\paragraph{Datasets.} We create three training datasets PUB, SYN, and MIX to study (1), (2), and (3), respectively. PUB consists of around 90k public examples including both samples from academic datasets and Web data (Common Crawl) labeled by our annotators. SYN adds additional 69k curated synthetic examples. MIX contains all examples in SYN with additional 60k production samples with labels. 

\paragraph{Models.} The baseline models are trained with basic supervised learning. The DAT models are trained with two hidden layers of 300 dimensions using additional 100k unlabeled production data points. All models are trained with up to 2 epochs, and the training is repeated 3 times with different random seeds.

\begin{table}[!t]
 \centering
 \resizebox{\columnwidth}{!}{

\begin{tabular}{lcc|cc|cc}
\toprule
     \textbf{Category} & \multicolumn{2}{c}{\textbf{PUB}} & \multicolumn{2}{c}{\textbf{SYN}} & \multicolumn{2}{c}{\textbf{MIX}} \\
     & Baseline & DAT & Baseline & DAT & Baseline & DAT \\
\midrule
    \w{S}    & .698 & \textbf{.730} & .726          & \textbf{.745}   & \textbf{.943} & .939 \\
    \w{H}      & .417 & \textbf{.491} & \textbf{.551} & .476            & \textbf{.843} & .818 \\
   \w{V}  & .490 & \textbf{.529} & \textbf{.532} & .531            &   \textbf{.640} & .633 \\
    \w{HR}  & .258 & \textbf{.369} & .326          & \textbf{.356}   & .453 & \textbf{.482} \\
    \w{SH}        & .063 & \textbf{.281} & .086          & \textbf{.296}   & .621 & \textbf{.632} \\
    \w{S3}        & .592 & \textbf{.759} & \textbf{.779} & .777 & .911     & \textbf{.936} \\
    \w{H2}      & .393 & \textbf{.643} & .570          & \textbf{.577}   & .851 & \textbf{.854} \\
    \w{V2}        & .165 & \textbf{.453} & .093          & \textbf{.507}   & .443 & \textbf{.533} \\
\bottomrule
\end{tabular}

}
 \caption{The average AUPRC on a production validation set. PUB denotes models trained on labeled public datasets, SYN adds additional synthetic examples, and MIX adds additional labeled production examples. We mark the best result within each configuration in \textbf{bold}.}
 \label{tb:dat_exp}
\end{table}

\paragraph{Results.} We compare the average AUPRC on the production validation set $\cV$. As demonstrated in \cref{tb:dat_exp}, the improvement from WDAT is significant when we only have access to public datasets (PUB), and the marginal gain reduces gradually as we add more training examples, especially in-distribution production samples.
For instance, DAT improved \w{SH} AUPRC from 0.063 to 0.281 on PUB and from 0.086 to 0.296 on SYN, whereas the improvement is only from 0.621 to 0.632 on MIX.
WDAT still helps weak categories (\w{SH} and \w{V2}) on SYN and MIX, but it may slightly hurt the performance for categories with a sufficient amount of in-distribution data such as \w{H} and \w{V}. We suspect this is because the model failed to find a representation that works very well for both the public datasets and our production distribution.
Further study on the model architecture and training methods is required to improve the performance on all categories with unlabeled data throughout different stages of the project.

\section{Related Work}

There is a long track record of work on the definition and detection of hateful, toxic, offensive and abusive content~\cite{kwok2013hate, nobata2016abusive, waseem2016racist, gibert2018hate, vidgen2019challenges, gehman2020realtoxicityprompts, rosenthal2020semisup, lees2022perspective}. \citet{zampieri2019toxic} proposed a three-level hierarchical taxonomy considering whether the given language is (i) offensive or not; (ii) targeted or not; and (iii) targeted at a group, an individual or other organizations. Usually hateful expressions targeting protected identity groups are considered hate speech~\cite{davidson2017automated}. Perspective API defines toxicity as "A rude, disrespectful, or unreasonable comment that is likely to make people leave a discussion". Some also used toxicity as a general umbrella term for offensive, abusive, and hateful language~\cite{pavlopoulos2020toxicity}. The definitions of hatefulness, toxicity, offensiveness and abusiveness have overlaps but are not exactly the same, creating obstacles for sharing datasets between projects. Furthermore, only a limited amount of work considered detailed subcategorizations~\cite{ethos, borkan2019civilcomments} to capture a spectrum of severity, making it harder to control labeling quality. Finally, there exist various types of potentially undesired text in the wild, such as sexual content involving minors, extreme graphic violence, or support for self-harm or suicides, besides offensive and abusive language, and we observed a gap between current research work and the entirety of content types that should be moderated and detected. 
Our work aims to fill in the gap.

Despite the common belief that training data quality is critical for model performance, there is still lack of community standards for labeling standards, annotator training, quality metrics, etc.~\cite{vidgen2020garbage, yin2021hatereview, lees2022perspective,pai2021}. \citet{vidgen2020garbage} studied 60+ datasets for abusive language detection and found that the primary data source is Twitter and expert coding is the most common way to annotate data, closely followed by crowdsourcing. For large-scale data collection, crowdsourcing remains the most common approach~\cite{ethos,zampieri2019toxic,davidson2017automated}. However, the weak skill set of non-expert annotators can lead to lower data quality~\cite{waseem2016racist,yin2021hatereview}. Some recent work turns to large pre-trained language models to generate synthetic data, significantly reducing the cost of time and human labor~\cite{wang2021gpt3help,hartvigsen2022toxigen}, but it is unclear whether model outputs would be diverse enough to adapt to the real-world distribution. Synthetic data can be hand-crafted~\cite{hatecheck}, but it is limited by size and thus more suitable for evaluation. It is noteworthy that training data can contain bias due to the subjectivity and biases in the data collection process \cite{davidson2019bias, sap2019bias}.

Active learning has been successfully applied to a number of different domains such as text classification \cite{lewis1994sequential,schohn2000less,siddhant-lipton-2018-deep}; machine translation \cite{zeng2019empirical}; image classification \cite{luo2005active,hoi2006batch,gal2017deep}; object detection \cite{schmidt2020advanced} and information retrieval \cite{shen2005active}. There are several families of active learning sampling strategies that are often used in practice. Uncertainty sampling selects data points about which the model is most uncertain. The uncertainty of the model can be quantified by predicted probabilities \cite{lewis1994sequential,lewis1994heterogeneous,culotta2005reducing,scheffer2001active}, disagreement among an ensemble of models \cite{seung1992query,dagan1995committee,mccallum1998em}, or by using dropout and Bayesian approaches \cite{gal2017deep, siddhant-lipton-2018-deep}. Diversity sampling chooses samples in a way that ensures sufficient diversity within the selection. This is commonly achieved by clustering unlabeled data and sampling from different clusters \cite{nguyen2004active,xu2007incorporating}, or by selecting samples which are "representative" of the sample distribution (i.e., which are similar to many other samples) \cite{mccallum1998em,settles2008analysis}. Uncertainty and diversity sampling are sometimes combined in a single complex active learning strategy.

Red-teaming is a common approach for model improvement by discovering and patching the weakness iteratively~\cite{dinan2019adv, vidgen2020worst, kiela2021dynabench, ziegler2022adv,perez2022red,ribeiro-etal-2020-beyond}, where humans are encouraged to look for examples that could fail the model. Dynabench~\cite{kiela2021dynabench} is built as a platform for easy adversarial data collection. \citet{mishkin2022risks} describes in detail an operational process for doing red-teaming using external experts. \citet{ziegler2022adv} designed a tool to efficiently assist human adversaries to identify failures in a classifier. Models trained with red-teaming data are found to be more robust to adversarial attack~\cite{dinan2019adv,ziegler2022adv} and human-in-the-loop dynamic data collection can efficiently improve model performance~\cite{kiela2021dynabench,vidgen2020worst}.

Domain adaptation aims at generalizing knowledge learned in the source domain towards a related target domain~\citep{BenDavid2006AnalysisOR, Weiss2016ASO, BenDavid2009ATO}, the technique is most useful when there is insufficient labeled data in the target domain but sufficient labeled data in the source domain. Different methods have been proposed to transfer the knowledge across domains~\citep{Ramponi2020NeuralUD, Blitzer2006DomainAW, Mansour2008DomainAW}.
Inspired by generative adversarial nets (GANs)~\citep{Goodfellow2014GenerativeAN} which train a discriminator to make the representations of source and target indistinguishable, Domain Adversarial Training (DAT) methods are proposed to reduce the domain discrepancy through a domain discriminator~\citep{pmlr-v70-arjovsky17a, Ganin2016DomainAdversarialTO, Tzeng2017AdversarialDD, Ganin2015UnsupervisedDA}. To learn domain-invariant feature representations, DAT employs a gradient reversal layer to maximize the minimal loss of the domain discriminator. However, DAT suffers from a gradient vanishing problem when the domain discriminator can tell apart the two domains easily, and Wasserstein distance based methods are proposed to enable a more stable training~\citep{Shen2018WassersteinDG, pmlr-v70-arjovsky17a, shah-etal-2018-adversarial}.

\section{Future Work and Limitations}
\label{sec:future}

\paragraph{Bias and Fairness.} Similar to other existing NLP models, our models also suffer from bias towards certain demographic attributes~\citep{Kusner2017Counterfactual, Garg2019Counterfactual, Dwork2012Fairness}. For instance, the model may give higher \w{hate} predictions if the input contains \w{gay} and higher \w{sexual} predictions if the input contains \w{her}. This is because we use data from the Internet, and social bias may present explicitly or implicitly in the training datasets. We tried mitigation methods such as creating a balanced synthetic dataset with templates but could not fully eliminate the issue. In the future, we will continue following related research and improve the fairness of our models.

\paragraph{Data Augmentation.} We plan to investigate more data augmentation methods to boost the training dataset. Although our current training dataset naturally includes misspelled words and incorrect grammar as some of it is user-generated content, it is valuable to experiment with data augmentation to improve lexicon robustness~\cite{wei2019eda,kobayashi2018contextual,zhang2021double} and the generalizability of the model~\cite{guo2019augmenting,shen2020cutoff,gao2021simcse}, especially when working with the changing distribution of real-world data. 

\paragraph{Better Multilingual Support.} Only about 5\% of the samples are non-English in our training set. As the vast majority of our production traffic is in English, we have not yet rigorously evaluated or optimized performance on non-English text. Multilingual toxic content classification~\cite{aluru2020deep,wang2021multilingual,lees2022perspective} would require more non-English training data and may need additional changes on tokenization or model architecture.

\paragraph{Red-teaming at scale.} Red-teaming is an effective way to find unknown failure cases for the model. Currently we do internal red-teaming with each new model version, which is not a scalable approach. In the future, we plan to set up a pipeline for model red-teaming similar to the one we have for labeling production traffic. We plan to use a specialized interface inspired by ~\citet{kiela2021dynabench,ziegler2022adv} to improve the efficiency of the red-teamers.

\paragraph{More Active Learning Experiments.} Our current active learning strategy to select high-value data for labeling is quite simple. For example, we did not explore diversity sampling due to computational restriction. Onward we plan to run more rigorous experiments comparing the performance of different active learning strategies, as well as more sophisticated strategies, incorporating both uncertainty and diversity sampling.

\section{Broader Impacts}

Content moderation classifiers have many uses. When paired with fair and robust enforcement practices, they have the potential to reduce certain instances of misuse \footnote{misuse may be defined as uses of the model that the moderating body does not want to allow, e.g. generation of hateful content} by ensuring that policies are operationalized on both inputs and outputs of language models. Classifiers also enable filtration of datasets at scale, which may be used to train language models with desired properties \cite{welbl2021challenges} and allow for better evaluation of language models \cite{gehman2020realtoxicityprompts}. Longer-term, content moderation classifiers can be used as a way to ensure high-stakes reliability in very-capable AI systems \cite{ziegler2022adv}---a critical necessity for enabling the deployment of those systems in certain domains.

While this underscores the importance of the undesired content classifiers, all classifiers rest on certain assumptions and decisions that may present vulnerabilities or make them inappropriate for certain use cases or types of text. Additionally, these tools can suffer from problematic biases, such as disproportionate false positives when discussing groups that are frequently the target of hate. \cite{Garg2019Counterfactual}

The following sections discuss the normative and subjective questions on which these classifiers rest and explore the challenges they present.




\subsection{Challenges of Taxonomy Design}

We take care to design our taxonomy to reflect generalizable viewpoints. However, much of our data is drawn from a US-centric context and the taxonomy was designed to best fit this data. Additionally, while we have designed our taxonomy to be as comprehensive as possible, it would still be useful for future researchers to add and update the categories based on their own use cases and deployment contexts. Given the sensitive nature of various tasks, we also encourage the use of this taxonomy in concert with other mitigation strategies, as there is no silver bullet for content moderation.

We hope that this work will encourage further discussion and debate around the principles and values that underpin content moderation.

\subsection{Annotator Viewpoints and Disagreement}

It is commonly agreed that the annotation of toxic language is subjective and that annotators' interpretations may be influenced by their personal and cultural backgrounds, including lived experiences, values and demographic factors. For example, \citet{waseem2016racist} found that feminist and anti-racist activists systematically disagree with crowd workers on their hate speech annotations. In their study, agreement between the authors, amateurs and expert annotators is low (14$\%$), most often because in many instances where the authors had identified hate speech, annotators do not. 


By necessity, incorporating diverse viewpoints invites disagreement on annotation labels. Much of the computer science literature focuses on eliminating inter-rater disagreements, most often via deliberation or majority vote. However, in the case of data from or about marginalized populations, disagreement may be a meaningful signal: An adverse effect of majority vote in such cases is limiting representation of minority perspectives in data \citet{prabhakaran2021}, potentially reinforcing societal disparities and harms. Moreover, analyzing disagreements may lead to a better understanding of the domain of application \citet{patton2019}.

In their study, rather than aggregating, \citet{Davani2021} preserve annotator disagreements, which they note could reflect useful and nuanced information about the uncertainty of a sample's membership to a class. Indeed, they demonstrate that their approach yields the same or better performance than similar approaches with aggregated labels, while retaining the ability to estimate uncertainty in predictions that correlate with real-life annotator disagreements.

Moreover, resolving disagreement via majority vote may be at odds with preserving minority opinions in subjective tasks. \citet{ovesdotter-alm-2011-subjective} argues that achieving a single real "ground truth" label is impossible and is not essential in subjective tasks, and calls for finding ways to model subjective interpretations of annotators, rather than seeking to reduce the variability in annotations.


\subsection{Annotator Selection and Welfare}

We are committed to ensuring that our labeling tasks are managed in a considerate and ethical manner, and we strive to follow current best practices for sourcing data labeling services \cite{pai2021}. Via our data vendors, all of our annotators are selected for their skill and willingness to participate in these difficult tasks. Before they opt in, all annotators are vetted by counselors and made aware of the risks and potential harms of working with sensitive data.  Our data vendors provide them with access to mental health and wellness resources and annotators have the right to opt out at any point.

\subsection{Data Privacy and Security}

Trustworthy handling of production data necessitates transparency with users and effective security measures. We obtain consent from all customers whose data is used to train our moderation models. Customers who wish to opt their data out of training may do so. No production data is included in the dataset we are releasing. Our data labeling and active learning pipelines feature security controls that are designed and tested to protect the confidentiality and integrity of production data. The model we deploy can not be used to generate text, only to compute safety scores, so we consider the risk of training data leakage to be extremely low.

\subsection{Summary of Broader Impacts Discussion}

Content moderation classifiers are one key tool that empowers developers of language models at every stage of the model development and deployment process- from working with large-scale datasets, to testing out models, to deploying the models to many users. However, as we have observed above, there are a range of normative and subjective decisions made throughout the development process of building these classifiers from designing taxonomies to labeling data. Given the nature of these tools, these decisions are sometimes distilled down bluntly and do not enable capturing the nuances that the moderation decision may warrant. This loss of nuance may disproportionately impact members of socially marginalized populations by muting their opinions via unweighted majority annotations. This impact is doubly grievous if moderation decisions about members of marginalized populations are made about them by a system that excludes their input. This highlights some inherent limitations of classifiers, using automated tools for content moderation, and point to the importance of their robust testing to ensure suitability for each specific use that they may be deployed in.

\section{Conclusion}

Building high-quality undesired content detection systems in the real world is a challenge that requires the incorporation of multiple methods. A good content taxonomy is the foundation for problem scoping and data collection. A reliable data pipeline is needed to guarantee high data quality and to handle distribution shift. We show that in cases where certain target content occurs rarely, an active learning sampling strategy leads to much better model performance. Additionally, we argue that good operational aspects of the labeling pipeline are essential for ensuring high data quality. And we show that model performance can further be improved through the use of curated synthetic data and semi-supervised learning.

As large generative language models become more and more prevalent, it becomes increasingly important to develop ways of controlling and guiding their outputs. The goal of this work has been to demonstrate one way of implementing such control by way of building content detection models. We are looking forward to further refinement of our approach in the future, as well as progress in other methods of controlling and aligning generative model outputs.

\section{Acknowledgments}

This work would not have been possible without the contributions of data workers. We greatly appreciate their work handling sensitive content and helping us build better automated systems to make content moderation work less demanding of human labor.

We also thank Miles Brundage, Raf Jakubanis, Gretchen Krueger, Derek Chen, Summer Yue, Karl Cobbe, Pranav Shyam, Jason Kwon and Matt Knight for feedback on this work.

\bibliography{anthology,custom}

\appendix








\section{Experiment Details}
\label{appendix:exp_details}

Table~\ref{ta:taxonomy_matching} presents how we map model taxonomies into labels of different evaluation datasets. Some of the mappings are only approximation. For example, Perspective defines "threat" as "Describes an intention to inflict pain, injury, or violence against an individual or group.", not including graphic violence, so not a perfect match for our "violence" category. Either or our taxonomy has a good match for "toxic", "severe toxic", or "offensive.

\define{Our Evaluation Set.} We are aware that about 4\% of our evaluation samples are in non-English. Perspective API call takes the language as an input parameter, but multilingual is not supported for several attributes. We instead use "en" for all the calls. 

\define{Jigsaw.} Jigsaw dataset is pretty large and we include about half of it into our training set to resolve the cold-start problem. Among the rest half, we sampled 5000 examples for evaluation.

\define{TweetEval.} We take the TweetEval~\cite{barbieri2020tweeteval} test datasets\footnote{\url{https://github.com/cardiffnlp/tweeteval/tree/main/datasets}} on "hate" and "offensive". There are in total 2970 samples in the hate task test set and 860 in the offensive one.

\define{Stormfront.} We use the test dataset of \citet{gibert2018hate}\footnote{\url{https://github.com/Vicomtech/hate-speech-dataset}}, containing 478 samples.

\define{Reddit.} We downsampled 5000 examples from the "RS\_201501" snapshot of Reddit pushshift datasets\footnote{\url{https://files.pushshift.io/reddit/submissions/}} and assigned noisy binary label to each example on whether it contains sexual content according to the subreddits as listed in \citet{barrientos2020erotic}.

\begin{table*}[t]
    \centering
    \resizebox{\textwidth}{!}{\begin{tabular}{ll|p{6.8cm}ll}
\toprule
\multicolumn{2}{c|}{Taxonomy} & Perspective & Ours \\
\midrule
Ours & Sexual & $\max$(sexually\_explicit, profanity, flirtation) & sexual \\
    & Hate & identity\_attack & hate \\
    & Violence & threat & violence \\
    & Harassment & max(toxicity, severe\_toxicity, insult, threat) & harassment \\
    & Sexual/minors & - & sexual/minors \\
\midrule
Jigsaw & Toxic & toxicity & harassment \\
    & Obscene & $\max$(sexually\_explicit, profanity) & sexual \\
    & Threat & threat & violence \\
    & Insult & insult & $\max$(harassment, hate) \\
    & Identity hate & identity\_attack & hate \\
\midrule
TweetEval & Hate & identity\_attack & hate \\
    & Offensive & $\max$(toxicity, severe\_toxicity, threat, insult, identity\_attack) & harassment \\
\midrule
Stormfront & Hate & identity\_attack & hate \\
\midrule
Reddit & Sexual & $\max$(sexually\_explicit, profanity, flirtation) & sexual \\
\bottomrule
\end{tabular}}
    \caption{How taxonomies of different APIs get mapped into labels of various evaluation datasets.}
    \label{ta:taxonomy_matching}
\end{table*}






\end{document}